# HIERARCHICAL DEEP LEARNING ARCHITECTURE FOR 10K OBJECTS CLASSIFICATION

Atul Laxman Katole[1], Krishna Prasad Yellapragada[1],
Amish Kumar Bedi[1], Sehaj Singh Kalra[1] and Mynepalli Siva Chaitanya[1]

[1] Samsung R&D Institute India - Bangalore, Bagmane Constellation Business Park, Doddanekundi Circle, Bangalore, India

## ABSTRACT

*Evolution of visual object recognition architectures based on Convolutional Neural Networks & Convolutional Deep Belief Networks paradigms has revolutionized artificial Vision Science. These architectures extract & learn the real world hierarchical visual features utilizing supervised & unsupervised learning approaches respectively. Both the approaches yet cannot scale up realistically to provide recognition for a very large number of objects as high as 10K. We propose a two level hierarchical deep learning architecture inspired by divide & conquer principle that decomposes the large scale recognition architecture into root & leaf level model architectures. Each of the root & leaf level models is trained exclusively to provide superior results than possible by any 1-level deep learning architecture prevalent today. The proposed architecture classifies objects in two steps. In the first step the root level model classifies the object in a high level category. In the second step, the leaf level recognition model for the recognized high level category is selected among all the leaf models. This leaf level model is presented with the same input object image which classifies it in a specific category. Also we propose a blend of leaf level models trained with either supervised or unsupervised learning approaches. Unsupervised learning is suitable whenever labelled data is scarce for the specific leaf level models. Currently the training of leaf level models is in progress; where we have trained 25 out of the total 47 leaf level models as of now. We have trained the leaf models with the best case top-5 error rate of 3.2% on the validation data set for the particular leaf models. Also we demonstrate that the validation error of the leaf level models saturates towards the above mentioned accuracy as the number of epochs are increased to more than sixty. The top-5 error rate for the entire two-level architecture needs to be computed in conjunction with the error rates of root & all the leaf models. The realization of this two level visual recognition architecture will greatly enhance the accuracy of the large scale object recognition scenarios demanded by the use cases as diverse as drone vision, augmented reality, retail, image search & retrieval, robotic navigation, targeted advertisements etc.*

## KEYWORDS

*Convolutional Neural Network [CNN], Convolutional Deep Belief Network [CDBN], Supervised & Unsupervised training*

## 1. INTRODUCTION

Deep learning based vision architectures learn to extract & represent visual features with model architectures that are composed of layers of non-linear transformations stacked on top of each other [1]. They learn high level abstractions from low level features extracted from images utilizing supervised or unsupervised learning algorithms. Recent advances in training CNNs with gradient descent based backpropagation algorithm have shown very accurate results due to inclusion of rectified linear units as nonlinear transformation [2]. Also extension of unsupervised learning algorithms that train deep belief networks towards training convolutional networks have exhibited promise to scale it to realistic image sizes [4]. Both the supervised and unsupervised learning approaches have matured and have provided architectures that can successfully classify objects in 1000 & 100 categories respectively. Yet both the approaches cannot be scaled realistically to classify objects from 10K categories.

The need for large scale object recognition is ever relevant today with the explosion of the number of individual objects that are supposed to be comprehended by artificial vision based solutions. This requirement is more pronounced in use case scenarios as drone vision, augmented reality, retail, image search & retrieval, industrial robotic navigation, targeted advertisements etc. The large scale object recognition will enable the recognition engines to cater to wider spectrum of object categories. Also the mission critical use cases demand higher level of accuracy simultaneously with the large scale of objects to be recognized.

In this paper, we propose a two level hierarchical deep learning architecture that achieves compelling results to classify objects in 10K categories. To the best of our knowledge the proposed method is the first attempt to classify 10K objects utilizing a two level hierarchical deep learning architecture. Also a blend of supervised & unsupervised learning based leaf level models is proposed to overcome labelled data scarcity problem. The proposed architecture provides us with the dual benefit in the form of providing the solution for large scale object recognition and at the same time achieving this challenge with greater accuracy than being possible with a 1-level deep learning architecture.

## 2. RELATED WORKS

We have not come across any work that uses 2-level hierarchical deep learning architecture to classify 10K objects in images. But object recognition on this large scale using shallow architectures utilizing SVMs is discussed in [5]. This effort presents a study of large scale categorization with more than 10K image classes using multi-scale spatial pyramids (SPM) [14] on bag of visual words (BOW) [13] for feature extraction & Support Vector Machines (SVM) for classification.

This work creates ten different datasets derived from ImageNet each with 200 to 10,000 categories. Based on these datasets it outlines the influence on classification accuracy due to different factors like number of labels in a dataset, density of the dataset and the hierarchy of labels in a dataset. The methods are proposed which provide extended information to the classifier on the relationship between different labels by defining a hierarchical cost. This cost is calculated as the height of the lowest common ancestor in WordNet. Classifiers trained on loss function using the hierarchical cost can learn to differentiate and predict between similar categories when compared to those trained on 0-1 loss. The error rate for the classification of entire 10K categories is not conclusively stated in this work.

## 3. PROBLEM STATEMENT

Supervised learning based deep visual recognition CNN architectures are composed of multiple convolutional stages stacked on top of each other to learn hierarchical visual features [1] as captured in Figure 1. Regularization approaches such as stochastic pooling, dropout, data augmentation have been utilized to enhance the recognition accuracy. Recently the faster convergence of these architectures is attributed to the inclusion of Rectified Linear Units [ReLU] nonlinearity into each of the layer with weights. The state of the art top 5 error rate reported is 4.9% for classification into 1K categories [6] that utilizes the above mentioned generic architectural elements in 22 layers with weights.

Unsupervised learning based architecture model as convolutional DBN learns the visual feature hierarchy by greedily training layer after layer. These architectures have reported accuracy of 65.4% for classifying 101 objects [4].

Both the architectures are not yet scaled for classification of 10K objects. We conjecture that scaling a single architecture is not realistic as the computations will get intractable if we utilize deeper architectures.

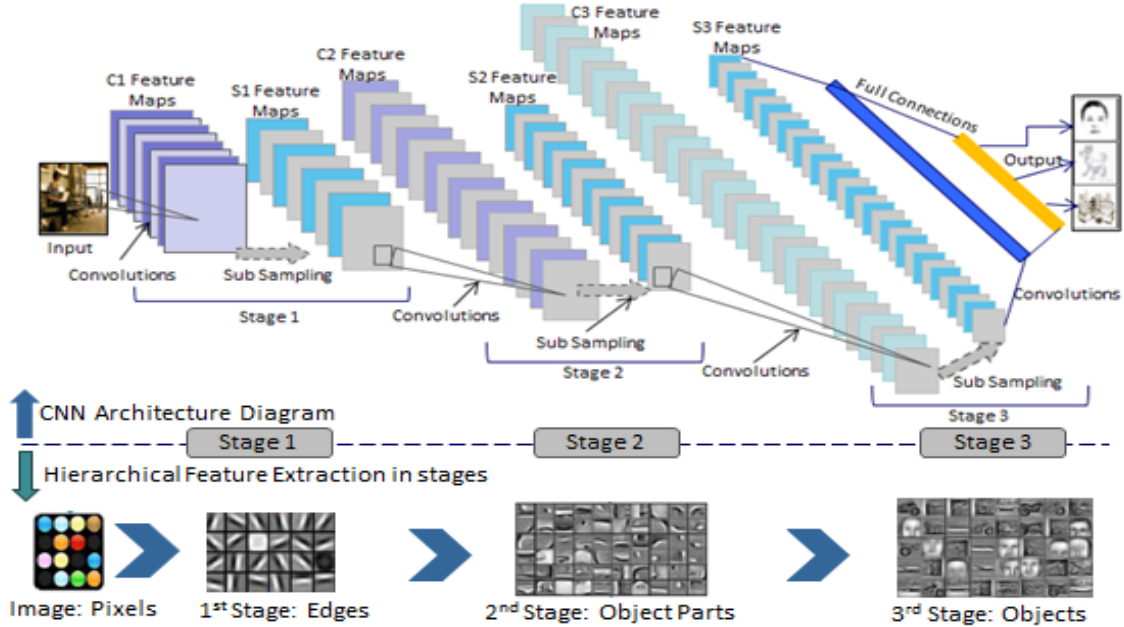

Figure 1. Learning hierarchy of visual features in CNN architecture

## 4. PROPOSED METHOD

We employ divide & conquer principle to decompose the 10K classification into root & leaf level distinct challenges. The proposed architecture classifies objects in two steps as captured below:

*1. Root Level Model Architecture:* In the first step the root i.e. the first level in architectural hierarchy recognizes high level categories. This very deep vision architectural model with 14 weight layers [3] is trained using stochastic gradient descent [2]. The architectural elements are captured in the table 1.

*2. Leaf Level Model Architecture:* In the second step, the leaf level recognition model for the recognized high level category is selected among all the leaf models. This leaf level model is presented with the same input object image which classifies it in a specific category. The leaf level architecture in the architectural hierarchy recognizes specific objects or finer categories. This model is trained using stochastic gradient descent [2]. The architectural elements are captured in the table 2.

CDBN based leaf level models can be trained with unsupervised learning approach in case of scarce labelled images [4]. This will deliver a blend of leaf models trained with supervised & unsupervised approaches. In all a root level model & 47 leaf level models need to be trained. We use ImageNet10K dataset [5], which is compiled from 10184 synsets of the Fall-2009 release of ImageNet. Each leaf node has at least 200 labelled images which amount to 9M images in total.

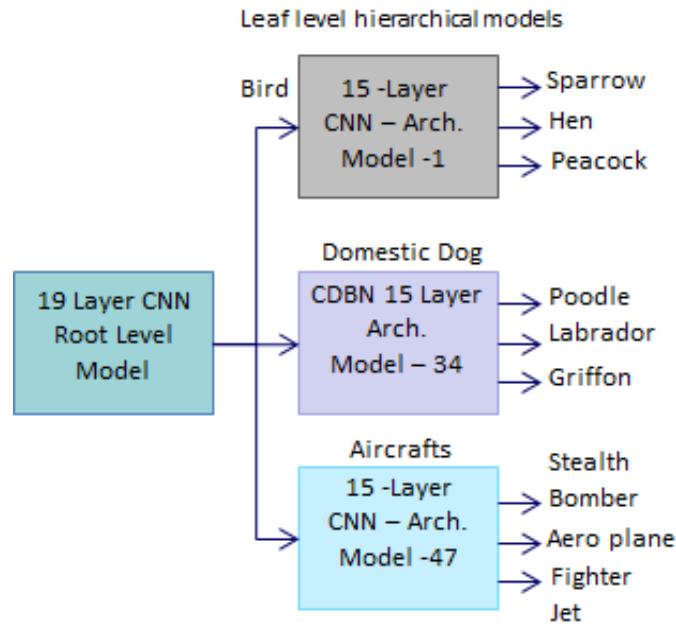

Figure 2. Two Levels – Hierarchical Deep Learning Archtecture

## 5. SUPERVISED TRAINING

In vision the low level features [e.g. pixels, edge-lets, etc.] are assembled to form high level abstractions [e.g. edges, motifs] and these higher level abstractions are in turn assembled to form further higher level abstractions [e.g. object parts, objects] and so on. Substantial number of the recognition models in our two-level hierarchical architecture is trained utilizing supervised training. The algorithms utilized for this method are referred to as error-back propagation algorithm. These algorithms require significantly high number of labelled training images per object category in its data set.

### 5.1. CNN based Architecture

CNN is a biologically inspired architecture where multiple trainable convolutional stages are stacked on the top of each other. Each CNN layer learns feature extractors in the visual feature hierarchy and attempts to mimic the human visual system feature hierarchy manifested in different areas of human brain as V1 & V2 [10]. Eventually the fully connected layers act as a feature classifier & learn to classify the extracted features by CNN layers into different categories or objects. The fully connected layers can be likened to the V4 area of the brain which classifies the hierarchical features as generated by area V2.

The root level & the leaf level CNN models in our architecture are trained with supervised gradient descent based backpropagation method. In this learning method, the cross entropy objective function is minimized with the error correction learning rule/mechanism. This mechanism computes the gradients for the weight updates of the hidden layers by recursively computing the local error gradient in terms of the error gradients of the next connected layer of neurons. By correcting the synaptic weights for all the free parameters in the network, eventually the actual response of the network is moved closer to the desired response in statistical sense.

Table 1. CNN architecture layers [L] with architectural elements for root level visual recognition architecture

| Layer No. | Type | Input Size | #kernels |
|---|---|---|---|
| 1 | Convolutional | 225 x 225 x 3 | 3 x 3 x 3 |
| 2 | Max Pool | 223 x 223 x 64 | 3 x 3 |
| 3 | Convolutional | 111 x 111 x 64 | 3 x 3 x 64 |
| 4 | Convolutional | 111 x 111 x 128 | 3 x 3 x 128 |
| 5 | Max Pool | 111 x 111 x 128 | 3 x 3 |
| 6 | Convolutional | 55 x 55 x 128 | 3 x 3 x 128 |
| 7 | Convolutional | 55 x 55 x 256 | 3 x 3 x 256 |
| 8 | Max Pool | 55 x 55 x 256 | 3 x 3 |
| 9 | Convolutional | 27 x 27 x 256 | 3 x 3 x 256 |
| 10 | Convolutional | 27 x 27 x 384 | 3 x 3 x 384 |
| 11 | Convolutional | 27 x 27 x 384 | 3 x 3 x 384 |
| 12 | Max Pool | 27 x 27 x 384 | 3 x 3 |
| 13 | Convolutional | 13 x 13 x 384 | 3 x 3 x 384 |
| 14 | Convolutional | 13 x 13 x 512 | 3 x 3 x 512 |
| 15 | Convolutional | 13 x 13 x 512 | 3 x 3 x 512 |
| 16 | Max Pool | 13 x 13 x 512 | 3 x 3 |
| 17 | Convolutional | 7 x 7 x 512 | 1 x 1 x 512 |
| 18 | Full-Connect | 12544 | 4096 |
| 19 | Full-Connect | 4096 | 2048 |
| 20 | Full-Connect | 2048 | 128 |
|  | 128-Softmax | 128 |  |

### 5.2. Architectural Elements

Architectural elements for the proposed architecture are:

- *Enhanced Discriminative Function*: We have chosen deeper architectures & smaller kernels for the root & leaf models as they make the objective function more discriminative. This can be interpreted as making the training procedure more difficult by making it to choose the feature extractors from higher dimensional feature space.

- *ReLU Non-linearity:* We have utilized ReLU nonlinearities as against sigmoidal i.e. non-saturating nonlinearities in each layer as it reduces the training time by converging upon the weights faster [2].

- *Pooling*: The output of convolutional-ReLU combination is fed to a pooling layer after alternative convolutional layers. The output of the pooling layer is invariant to the small changes in location of the features in the object. The pooling method used is either max-pooling OR stochastic pooling. Max Pooling method averages the output over the neighborhood of the neurons where-in the poling neighborhoods can be overlapping or non-overlapping. In majority of the leaf models we have used Max Pooling approach with overlapping neighborhoods.

   Alternatively we have also used Stochastic Pooling method when training for few models. In Stochastic Pooling the output activations of the pooling region are randomly picked from the activations within each pooling region, following multinomial distribution. This distribution is computed from the neuron activation within the given region [12]. This approach is hyper parameter free. The CNN architecture for stochastic pooling technique is captured in table 3.

- Dropout: With this method the output of each neuron in fully connected layer is set to zero with probability 0.5. This ensures that the network samples a different architecture when a new training example is presented to it. Besides this method enforces the neurons to learn more robust features as it cannot rely on the existence of the neighboring neurons.

Table 2. CNN architecture layers [L] with architectural elements for leaf level visual recognition architecture with Max pooling strategy

| Layer No. | Type | Input Size | #kernels |
|---|---|---|---|
| 1 | Convolutional | 225 x 225 x 3 | 7 x 7 x 3 |
| 2 | Max Pool | 111 x 111 x 64 | 3 x 3 |
| 3 | Convolutional | 55 x 55 x 64 | 3 x 3 x 64 |
| 4 | Convolutional | 55 x 55 x 128 | 3 x 3 x 128 |
| 5 | Max Pool | 55 x 55 x 128 | 3 x 3 |
| 6 | Convolutional | 27 x 27 x 128 | 3 x 3 x 128 |
| 7 | Convolutional | 27 x 27 x 256 | 3 x 3 x 256 |
| 8 | Max Pool | 27 x 27 x 256 | 3 x 3 |
| 9 | Convolutional | 13 x 13 x 256 | 3 x 3 x 256 |
| 10 | Convolutional | 13 x 13 x 384 | 3 x 3 x 384 |
| 12 | Max Pool | 13 x 13 x 384 | 3 x 3 |
| 13 | Convolutional | 7 x 7 x 384 | 1 x 1 x 384 |
| 14 | Full-Connect | 6272 | 2048 |
| 15 | Full-Connect | 2048 | 2048 |
| 16 | Full-Connect | 2048 | 256 |
|  | 256-Softmax | 256 |  |

Table 3. CNN architecture layers [L] with architectural elements for leaf level visual recognition architecture with stochastic pooling strategy

| Layer No. | Type | Input Size | #kernels |
|---|---|---|---|
| 1 | Convolutional | 225 x 225 x 3 | 7 x 7 x 3 |
| 2 | Stochastic Pool | 111 x 111 x 64 | 3 x 3 |
| 3 | Convolutional | 55 x 55 x 64 | 3 x 3 x 64 |
| 4 | Convolutional | 55 x 55 x 128 | 3 x 3 x 128 |
| 5 | Stochastic Pool | 55 x 55 x 128 | 3 x 3 |
| 6 | Convolutional | 27 x 27 x 128 | 3 x 3 x 128 |
| 7 | Convolutional | 27 x 27 x 256 | 3 x 3 x 256 |
| 8 | Stochastic Pool | 27 x 27 x 256 | 3 x 3 |
| 9 | Convolutional | 13 x 13 x 256 | 3 x 3 x 256 |
| 10 | Convolutional | 13 x 13 x 384 | 3 x 3 x 384 |
| 12 | Stochastic Pool | 13 x 13 x 384 | 3 x 3 |
| 13 | Convolutional | 7 x 7 x 384 | 1 x 1 x 384 |
| 14 | Full-Connect | 6272 | 2048 |
| 15 | Full-Connect | 2048 | 2048 |
| 16 | Full-Connect | 2048 | 256 |
|  | 256-Softmax | 256 |  |

### 5.3. Training Process

We modified libCCV open source CNN implementation to realize the proposed architecture which is trained on NVIDIA GTX™ TITAN GPUs. The root & leaf level models are trained using stochastic gradient descent [2].

The leaf models are trained as batches of 10 models per GPU on the two GPU systems simultaneously. The first 4 leaf models were initialized and trained from scratch for 15 epochs with learning rate of 0.01, and momentum of 0.9. The rest of the leaf models are initialized from the trained leaf models and trained with learning rate as 0.001. The root model has been trained for 32 epochs with learning rate of 0.001, after having been initialized from a similar model trained on ImageNet 1K dataset.

It takes 10 days for a batch of 20 leaf models to train for 15 epochs. Currently the root model and 25/47 leaf models have been trained in 5 weeks. Full realization of this architecture is in progress and is estimated to conclude by second week of September'15.

## 6. UNSUPERVISED TRAINING

Statistical Mechanics has inspired the concept of unsupervised training fundamentally. Specifically statistical mechanics forms the study of macroscopic equilibrium properties of large system of elements starting from the motion of atoms and electrons. The enormous degree of freedom as necessitated by statistical mechanics foundation makes the use of probabilistic methods to be the most suitable candidate for modelling features that compose the training data sets [9].

### 6.1. CDBN based Architecture

The networks trained with statistical mechanics fundamentals model the underlying training dataset utilizing Boltzmann distribution. To obviate the painfully slow training time as required to train the Boltzmann machines, multiple variants of the same have been proposed where the Restricted Boltzmann Machine [RBM] is the one that has provided the best possible modelling capabilities in minimal time. The resulting stacks of RBM layers are greedily trained layer by layer [4] resulting in the Deep Belief Networks [DBN] that successfully provides the solution to image [1- 4], speech recognition [8] and document retrieval problem domains.

DBN can be described as multilayer generative models that learn hierarchy of non-linear feature detectors. The lower layer learns lower level features which feeds into the higher level and help them learn complex features. The resulting network maximizes the probability that the training data is generated by the network. But DBN has its own limitations when scaling to realistic image sizes [4]. First difficulty is to be computationally tractable with increasing image sizes. The second difficulty is faced with lack of translational invariance when modelling images.

To scale DBN for modelling realistic size images the powerful concept of Convolutional DBN [CDBN] had been introduced. CDBN learns feature detectors that are translation invariant i.e. the feature detectors can detect the features that can be located at any location in an image.

We perform the block Gibbs sampling using conditional distribution as suggested in [4] to learn the convolutional weights connecting the visible and hidden layers where v and h are activations of neurons in visible & hidden layers respectively. Also $b_j$ are hidden unit biases and $c_i$ are visible unit biases. W forms the weight matrix connecting the visible and hidden layer. The Gibbs sampling is conducted utilizing (1) & (2).

$$P(h_{ij} = 1 \mid v) = sigmoid((W * v)_{ij} + b_j) \quad (1)$$

$$P(v_{ij} = 1 \mid h) = sigmoid((\sum W * v)_{ij} + c_i) \quad (2)$$

The weights such learnt give us the layers for Convolutional RBMs [CRBM]. The CRBMs can be stacked on top of each other to form CDBN. We had probabilistic Max Pooling layers after convolutional layers [4]. MLP is introduced at the top to complete the architecture. This concept is captured in Figure 3.

We train the first two layers in the leaf architecture with unsupervised learning. Later we abandon unsupervised learning and use the learnt weights in the first two layers to initialize the weights in CNN architecture. The CNN architecture weights are then fine-tuned using backpropagation method. The architecture used for training with unsupervised learning mechanism is same as captured in table 2. Also a two-level hierarchical deep learning architecture can be constructed entirely with CDBN as depicted in Figure 4.

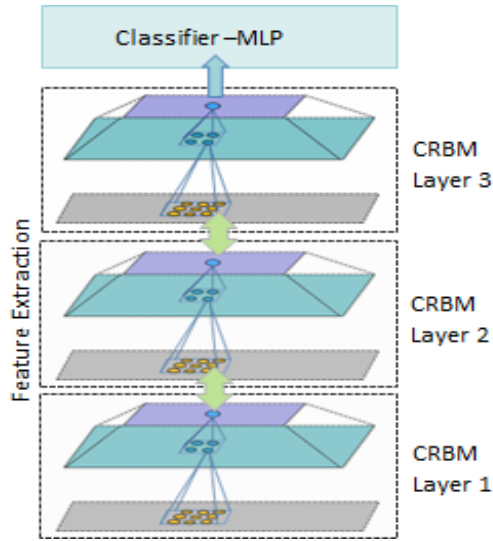

Figure 3. Convolutional DBN constructed by stacking CRBMs which are learnt one by one.

### 6.2. Training Process

We have used Contrastive Divergence CD-1 mechanism to train the first two layers of the architecture as specified for unsupervised learning. The updates to the hidden units in the positive phase of CD-1 step were done with sampling rather than using the real valued probabilities. Also we had used mini-batch size of 10 when training.

We had monitored the accuracy of the training utilizing -
1. Reconstruction Error: It refers to the squared error between the original data and the reconstructed data. While it does not guarantee accurate training, but during the course of training it should generally decrease. Also, any large amount of increase suggests the training is going wrong.

2. Printing learned Weights: The learned weighs needs to be eventually visualized as oriented, localized edge filters. Printing weights during training helps identify whether the weights are "approaching" that filter-like shape.

When the ratio of variance of reconstructed data to variance of input image exceeds 2, we decrease the learning rate by factor of 0.9 and reset the values of weights and hidden bias updates to ensure that weights don't explode. The initial momentum was chosen to be 0.5 which is increased finally to 0.9. The initial learning rate is chosen to be 0.1.

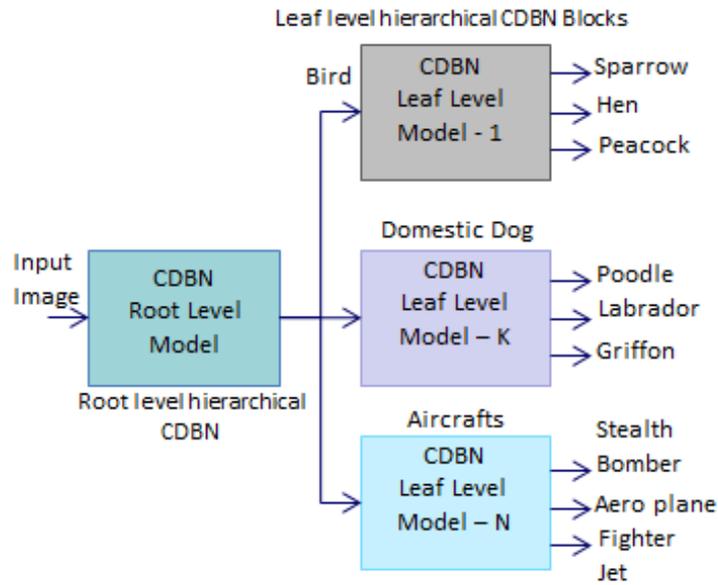

Figure 4. Proposed 2 Level Hierarchical Deep Learning Architecture constructed entirely utilizing CDBNs for classification/Recognition of 10K+ objects

## 7. TWO-LEVEL HIERARCHY CREATION

The 2-level hierarchy design for classification of 10K objects categories requires decision making on the following parameters –
1. Number of leaf models to be trained and
2. Number of output nodes in each leaf model.

To decide these parameters, we first build a hierarchy tree out of the 10184 synsets (classes) in ImageNet10K dataset (as described in section 7.1). Then using a set of thumb-rules (described in section 7.2), we try to split and organize all the classes into 64 leaf models, each holding a maximum of 256 classes.

### 7.1. Building Hierarchical Tree

Using the WordNet IS A relationship, all the synsets of ImageNet10K dataset are organized into a hierarchical tree. The WordNet ISA relationship is a file that lists the parent-child relationships between synsets in ImageNet. For example a line "n02403454 n02403740" in the relationship file refers to the parent synset to be n02403454 (cow) and child synset as n02403740 (heifer). However the ISA file can relate a single child to multiple parents, i.e. heifer is also the child of another category n01321854 (young mammal). As the depth of a synset in ImageNet hierarchy has no relationship to its semantic label, we focused on building the deepest branch for a synset. We utilized a simplified method that exploits the relationship between synset ID and depth; the deeper a category *nXXX* the larger its number *XXX*. Hence we used the parent category of *heifer* as *cow*, instead of *young mammal*.

The algorithm Htree as depicted in Figure 5a & 5b is used to generate the hierarchy tree. In this paper, the results from the *ninth (9th) iteration* are used as base tree. A sample illustration is captured in Figure 6.

### 7.2. Thumb-rules for Building Hierarchical Tree

From the Hierarchy tree, it is evident that the dataset is skewed towards the categories like flora, animal, fungus, natural objects, instruments etc. that are at levels closer to the tree root i.e. 80% of the synsets fall under 20% of the branches.

```
Algorithm: deepest_branch
Finding the deepest branch for a synset
```
- Input:
    - synset
    - WordNet_ISA.map
- parents := WordNet_ISA.map[synset]
- closest_parent := Max([XXX from nXXX of parents])
- branch := deepest_branch(closest_parent, WordNet_ISA.map)
- append closest_parent to branch
- return branch

Figure 5a. Pseudo-code for Hierarchy Tree Generation (HTree) Algorithm

```
Algorithm: Generate Hierarchy Tree for the Synsets in ImageNet10K, using deepest_branch algorithm.
```
- Input:
    - Synsets.list
    - WordNet_ISA.map
- branches := []
- for each synset in synset.list
    - branch := deepest_branch(synset, WordNet_ISA.map)
    - append branch to branches
- depths := [len(branch) for branch in branches]
- max_depth := max(depths)
- for iter := [max_depth .. 1]
    - roll-up and merge leafs where depth = iter

Figure 5b. Pseudo-code for Hierarchy Tree Generation (HTree) Algorithm

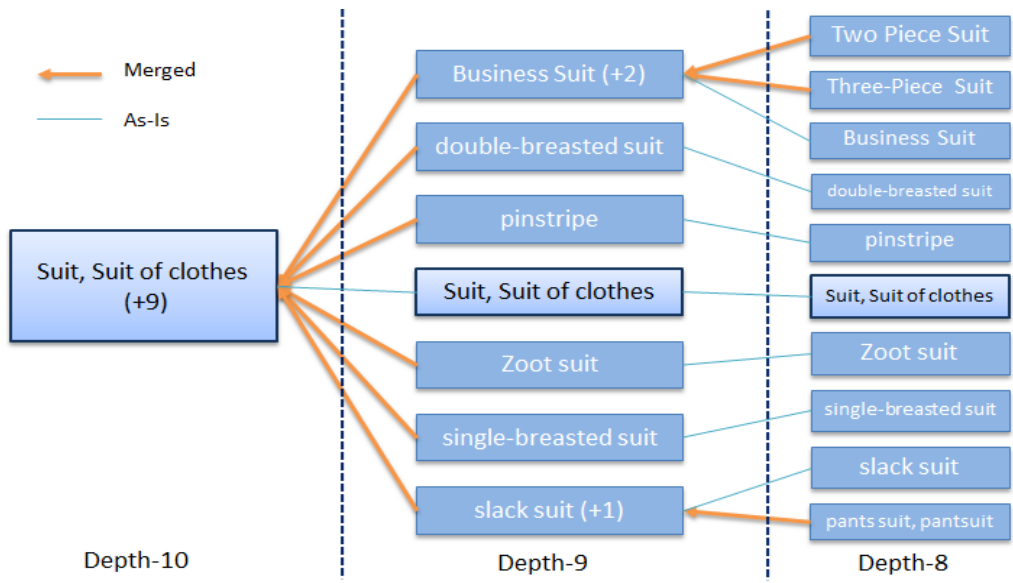

Figure 6. Hierarchy Tree Generated by HTree algorithm at iterations 8, 9 & 10 for ImageNet10K. The categories at Depth-9 are part of leaf-1 model.

Taking into account the number of models to be trained and the time & resources required for fine-tuning each model, the below thumb-rules were decided to finalize the solution parameters:
1. The ideal hierarchy will have 64 leaf models each capable of classifying 256 categories
2. The synsets for root and leaf level models have to be decided such that the hierarchy tree is as flat as possible
3. The total number of synsets in a leaf model should be less than 256 and
4. If leaf level models have more than 256 sub-categories under it, the remaining subcategories will be split or merged with another leaf.

### 7.3. Final Solution Parameters

The final solution parameters are as follows, 47 leaf models and with each leaf model classifying 200 ~ 256 synsets.

### 8. RESULTS

We formulate the root & leaf models into 2-level hierarchy. In all a root level model & 47 leaf level models need to be trained. Each leaf level model recognizes categories that range from 45 to 256 in numbers. We use ImageNet10K dataset [7], which is compiled from 10184 synsets of the Fall-2009 release of ImageNet.

Each leaf node has at least 200 labelled images which amount to 9M images in total. The top 5 error rates for 25 out of 47 leaf level models have been computed. The graph in Fig. 5 plots the top 5 errors of leaf models vis-à-vis the training epochs. We observe that when the leaf models are trained with higher number of training epochs the top 5 error decreases. The top 5 error rate for the complete 10K objects classification can be computed upon training of all the 47 models as required by the 2-level hierarchical deep learning architecture.

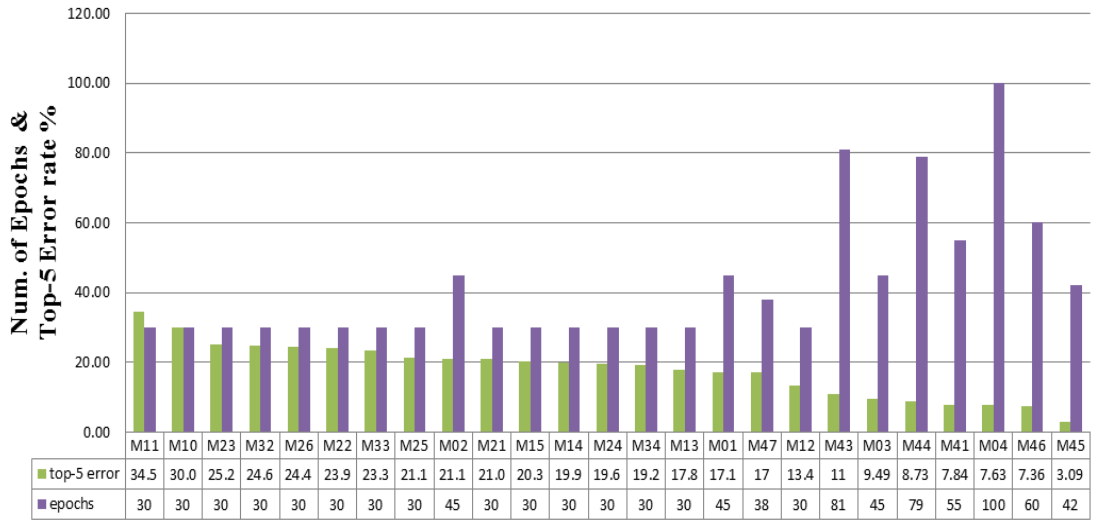

Figure 7. Graph captures the decrease in top-5 error rate with increase in increase in number of epochs when training with supervised training method

Training for classification of 10K objects with the proposed 2-level hierarchical architecture is in progress and is estimated to be completed by mid of September'15.

In this architecture, the root model & the 46/47 leaf models are based on CNN architecture and trained with supervised gradient descent.

Utilizing unsupervised learning we have trained a leaf model Leaf-4 that consists of man-made artifacts with 235 categories. The model for Leaf-4 is CDBN based architecture as described in Section 6. We have trained the first layer of this CDBN architecture with contrastive divergence (CD-1) algorithm. Later on the first layer weights are utilized to initialize the weights for leaf-4 model in supervised setting. The same are then fine-tuned with back-propagation utilizing supervised learning. The feature extractors or kernels learnt with supervised & unsupervised learning are captured in Fig. 7.

We intend to compute the top 5 error rates for categories using the algorithm as captured in the Fig. 9. Figures 10 – 13 depict the Top-5 classification results using this 2-level hierarchical deep learning architecture. From top-left, the first image is the original image used for testing. The remaining images represent the top-5 categories predicted by the hierarchical model, in descending order of their confidence.

## 9. CONCLUSIONS

The top-5 error rate for the entire two-level architecture is required to be computed in conjunction with the error rates of root & leaf models. The realizations of this two level visual recognition architecture will greatly simply the object recognition scenario for large scale object recognition problem. At the same time the proposed two level hierarchical deep learning architecture will help enhance the accuracy of the complex object recognition scenarios significantly that otherwise would not be possible with just 1-level architecture.

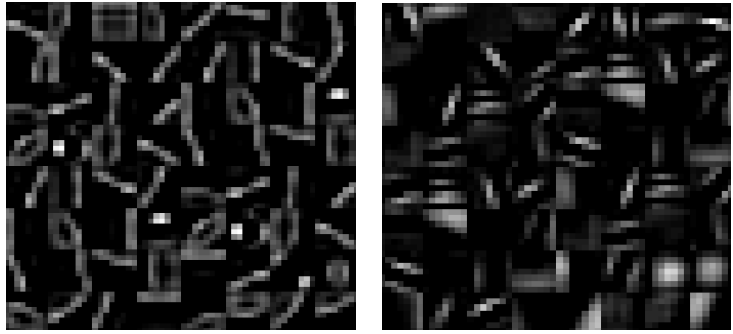

Figure 8. : The left image depicts the first layer 64 numbers of 7*7 filter weights learned by leaf-4 utilizing CD-1. The right image depicts the same first layer weights after fine-tuning with back propagation.

The trade of with the proposed 2-level architecture is the size of the hierarchical recognition model. The total size of the 2-level recognition models including the root & leaf models amounts to approximately 5 GB. This size might put constraints towards executing the entire architecture on low-end devices. The same size is not a constraint when executed with high end device or cloud based recognition where RAM size is higher. Besides we can always split the 2-level hierarchical model between device & cloud which paves the way for object recognition utilizing the novel device-cloud collaboration architectures.

The proposed 10K architecture will soon be available for classifying large scale objects. This breakthrough will help enable applications as diverse as drone vision, industrial robotic vision, targeted advertisements, augmented reality, retail, robotic navigation, video surveillance, search & information retrieval from multimedia content etc. The hierarchical recognition models can be deployed and commercialized in various devices like Smart phones, TV, Home Gateways and VR head set for various B2B and B2C use cases.

---

Algorithm 1: Error eval. algorithm for 10K 2-level Hierarchical architecture

- Input:
  - ROOT.model, LEAF[1-N].model, LID_GID.map
  - ROOT Test Data: {GID, RID, X}$^+$, where
    - GID is the global synset ID [1-10184]
    - RID is the root category [1-47]
    - X is the input image
- top5_error = 0
- For each test in {GID, RID, X}$^+$
  - rootTop5 := classify(ROOT.model, X) where rootTop5: {RID, RCONF}$^5$
  - For each RID in rootTop5:
    - leafTop5[RID] := classify(LEAF[RID].model, X) where leafTop5: {LID, LCONF}$^5$
    - leafTop5[RID].LCONF *= rootTop5.RCONF
  - sort leafTop5 in DESC
  - Map each LID in leafTop5 to GID using LID_GID.map
  - If test.GID **not in** leafTop5
    - Increment top5_error
- COMPUTE top5_error in %tage

---

Figure 9. Algorithm to compute top 5 error rates for 10K categories as evaluated by 2-level hierarchical deep learning algorithm

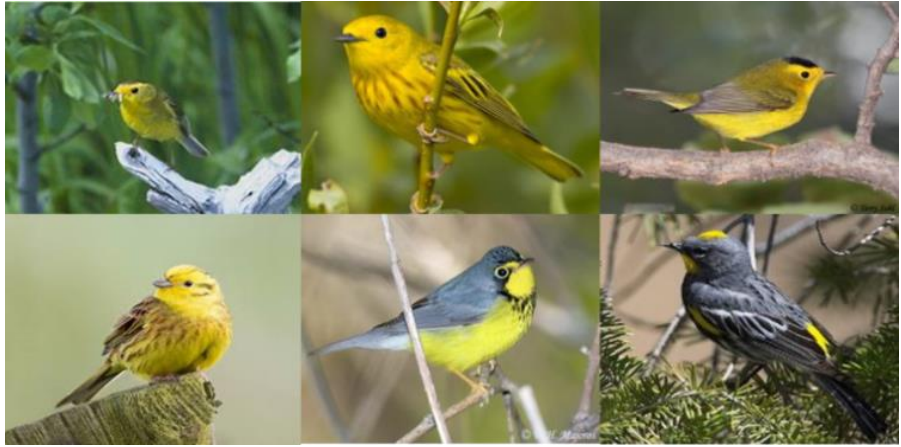

Figure 10. The test image belongs to Wilson's warbler. The predicted categories in order of confidence are a) Yellow Warbler b) Wilsons Warbler c) Yellow Hammer d) Wren, Warbler e) Audubon's Warbler

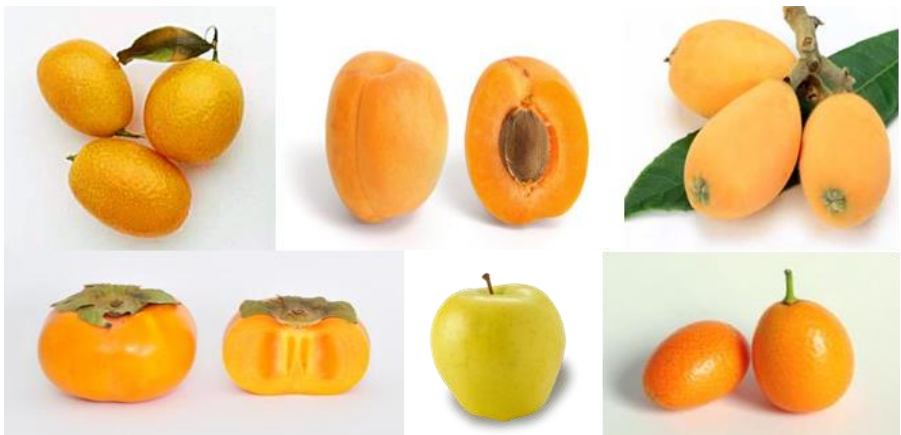

Figure 11. The test image belongs to fruit Kumquat. The predicted categories in order of confidence are a) Apricot b) Loquat c) Fuyu Persimmon d) Golder Delicious e) Kumquat

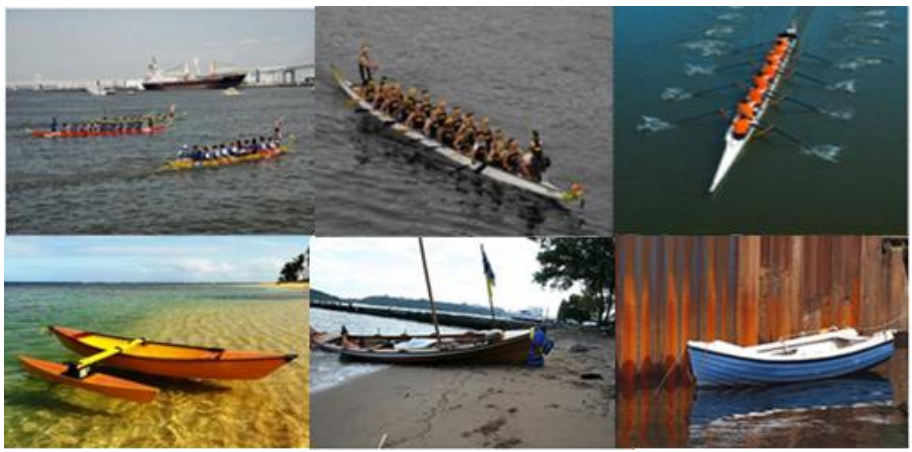

Figure 12. The test image belongs to Racing Boat. The top-5 predicted categories are a) Racing Boat b) Racing shell c) Outrigger Canoe d) Gig e) Rowing boat

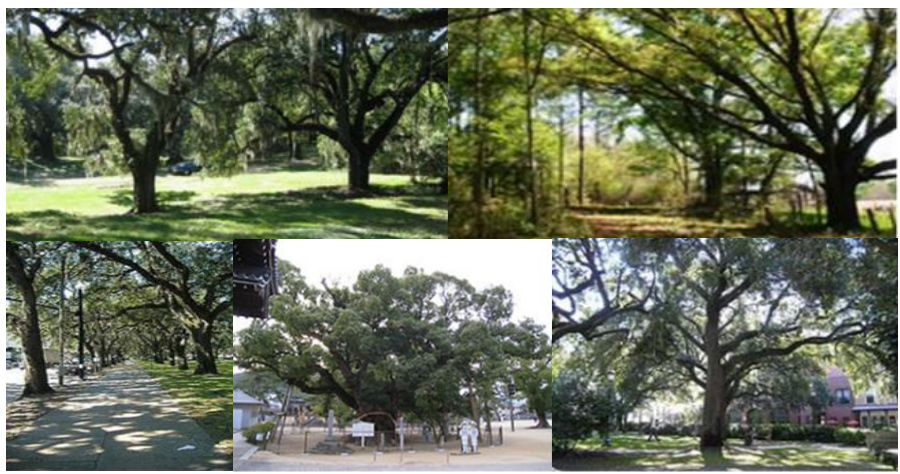

Figure 13. The test image belongs to category Oak. The top-5 predicted categories are a) Live Oak b) Shade Tree c) Camphor d) Spanish Oak

## ACKNOWLEDGEMENTS

We take this opportunity to express gratitude and deep regards to our mentor Dr. Shankar M Venkatesan for his guidance and constant encouragement. Without his support it would not have been possible to materialize this paper.

## AUTHORS

**Atul Laxman Katole**

Atul Laxman Katole has completed his M.E in Jan 2003 in Signal Processing from Indian Institute of Science, Bangalore and is currently working at Samsung R&D Institute India-Bangalore. His technical areas of interest include Artificial Intelligence, Deep Learning, Object Recognition, Speech Recognition, Application Development and Algorithms. He has seeded & established teams in the domain of Deep Learning with applications to Image & Speech Recognition.

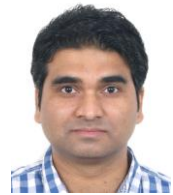

**Krishna Prasad Yellapragada**

Krishna Prasad Yellapragada is currently working as a Technical Lead at Samsung R&D Institute India-Bangalore. His interests include Deep Learning, Machine Learning & Content Centric Networks.

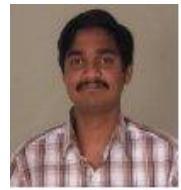

**Amish Kumar Bedi**

Amish Kumar Bedi has completed his B.Tech in Computer Science from IIT Roorkee, 2014 Batch. He is working in Samsung R&D Institute India-Bangalore' since July 2014. His technical areas of interest include Deep Learning/Machine Learning and Artificial Intelligence.

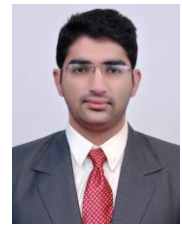

**Sehaj Singh Kalra**

Sehaj Sing Kalra has completed his B.Tech in Computer Science from IIT Delhi, batch of 2014. He is working in Samsung R&D Institute India - Bangalore since June 2014. His interest lies in machine learning and its applications in various domains, specifically speech and image.

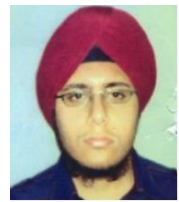

**Mynepalli Siva Chaitanya**

Mynepalli Siva Chaitanya has completed his B.Tech in Electrical Engineering from IIT Bombay, batch of 2014. He is working in Samsung R&D Institute India - Bangalore since June 2014. His area of interest includes Neural Networks and Artificial Intelligence.

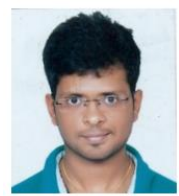